\pdfoutput=1

\documentclass[11pt]{article}

\usepackage[preprint]{acl}

\usepackage{times}
\usepackage{latexsym}

\usepackage[T1]{fontenc}

\usepackage[utf8]{inputenc}

\usepackage{microtype}

\usepackage{inconsolata}

\usepackage{graphicx}

\usepackage{bigstrut,multirow,rotating,booktabs}
\usepackage{tabularx}
\title{MTMT: Consolidating Multiple Thinking Modes to Form a Thought Tree for Strengthening LLM}

\author{Changcheng Li \And Xiangyu Wang \And Qiuju Chen \\
  University of Science and Technology of China \\
\And Xiren Zhou \And Huanhuan Chen 
}

\begin{document}
\maketitle
\begin{abstract}
Large language models (LLMs) have shown limitations in tasks requiring complex logical reasoning and multi-step problem-solving. To address these challenges, researchers have employed carefully designed prompts and flowcharts, simulating human cognitive processes to enhance LLM performance, such as the Chain of Thought approach. In this paper, we introduce \textbf{MTMT} (Multi-thinking Modes Tree), a novel method that interacts with LLMs to construct a thought tree, simulating various advanced cognitive processes, including but not limited to association, counterfactual thinking, task decomposition, and comparison. By breaking down the original complex task into simpler sub-questions, MTMT facilitates easier problem-solving for LLMs, enabling more effective utilization of the latent knowledge within LLMs. We evaluate the performance of MTMT under different parameter configurations, using \textbf{GPT-4o mini} as the base model. Our results demonstrate that integrating multiple modes of thinking significantly enhances the ability of LLMs to handle complex tasks.
\end{abstract}

\section{Introduction}
With the development of natural language processing, large language models (LLMs) have come to play a pivotal role in the field. There is hope that these models can solve most natural language problems and even achieve Artificial General Intelligence (AGI). However, despite increasing the amount of data and model parameters, the natural language capabilities of large models have not achieved the previously astonishing breakthroughs. Since the release of GPT-4 \citep{openai2023gpt}, other large language models may have achieved better results in some areas compared to GPT-4, but there has been no significant change \citep{Claude,gemini}. Additionally, issues such as model hallucinations, unfaithful explanations, lack of memory, and inadequate logical reasoning continue to challenge researchers.

To address these issues, many contextual learning schemes have been proposed to improve the performance of large models in various domains. For instance, algorithms like Chain of Thought \citep{wei2022chain} and many other algorithms, similar to Chain of Thought methods, enhance model output accuracy by decomposing tasks and employing a series of evaluation and retrospection techniques. Sys2 attention \citep{weston2023system} improves outcomes by asking about key segments in the prompts given to the LLM, thereby strengthening focus on important content. Additionally, methods such as the work of \citet{ma2023let} use counterfactuals to enhance the LLM's ethical level.

All the aforementioned methods are very similar to the thinking patterns people use when solving complex problems. Psychology suggests that humans have two modes of thinking: System 1 and System 2 \citep{thinking}. Inspired by neuroscience and psychology \citep{daw2005uncertainty,thinking}, we believe that the characteristics exhibited by today's LLMs are very similar to System 1, which is highly intuitive and relies on probabilities. As defined in the work of \citet{yu2024distilling}, we define System 1 reasoning as the immediate responses generated by the LLM for the given problems. System 2 reasoning refers to any method that involves producing intermediate tokens, such as approaches that perform searches or use multiple prompts before delivering a final answer.

\begin{figure*}[t]
    \centering
    \includegraphics[width=0.93\textwidth,trim={0.5cm 0.5cm 0.2cm 0.5cm},clip]{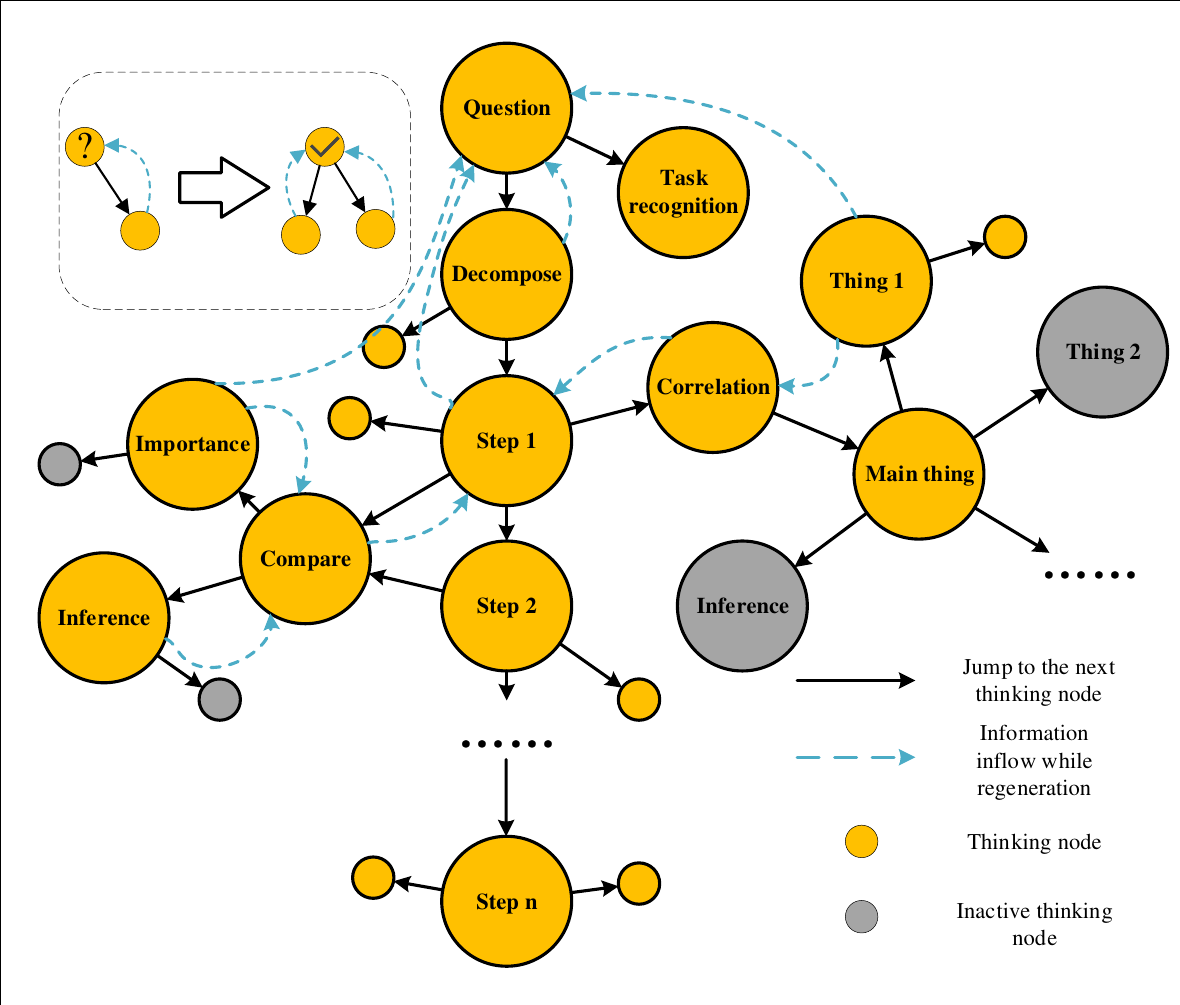}
    \caption{The overall structure of MTMT. Starting from the main question, nodes that are "uncertain" about their own answers will generate different sub-nodes. Relevant information is passed to ancestor nodes, while irrelevant or faulty nodes are marked as deactivated.}
    \label{net_diagram}
\end{figure*}

Building on this foundation, we designed an algorithm called MTMT (Multi-thinking Modes Tree), a graph that interacts repeatedly with the model to achieve the effects of System 2. As shown in Figure \ref{net_diagram}, the MTMT algorithm integrates various thinking modes to enhance the logical reasoning capabilities and accuracy of answers provided by LLMs. By thinking around the original complex task, we derive many simpler sub-questions. Through repeated interactions with the LLM, we aim to extract all the potential knowledge and solution strategies stored within it regarding this issue, thereby achieving the best performance of the LLM in a zero-shot setting.

\begin{figure}[t]
    \centering
    \includegraphics[width=0.45\textwidth]{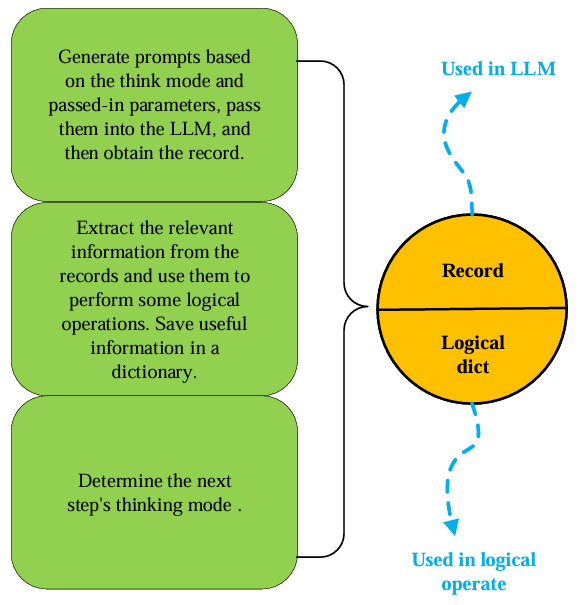}
    \caption{Flowchart of each thinking node. Generate and answer questions, extract information, decide the next thinking mode, and transition to or generate a thinking node.}
    \label{framework}
\end{figure}

The MTMT employs a state machine-like approach, based on the current state, we select a thinking mode (e.g., decomposition, comparison, analogy, reasoning and so on) and generate the corresponding prompt. The results from each thinking mode are stored as nodes in a graph. Next, we process the existing information and finally determines the next strategy to use and whether to continue or stop. The structural diagram of MTMT and the node flowchart are shown in Figures \ref{net_diagram} and \ref{framework}.

We will demonstrate the algorithm's effectiveness on the GPQA, TruthfulQA, and GSM8K datasets. On the GPQA dataset, we achieved an accuracy of 44.0\% without using external knowledge. Specific results can be found in Table \ref{table}. In fact, as a general strategy, our approach can achieve good results across various datasets without the need to adjust prompts for different datasets.

The main contributions of this paper are as follows:
\begin{itemize}
\item A System 2 reasoning method integrating multiple thinking modes is proposed. These diverse thinking modes enable large models to think more comprehensively and deeply, improving accuracy.
\item  Through the thinking graph, we can capture the entire reasoning process of the LLM, making it easier to trace the source of errors, which enhances the interpretability of the answers.
\item By interacting with the large model in various ways, we can uncover as much hidden potential knowledge about the problem within the model, thereby enhancing its performance.
\item As a generalizable thinking method, it can be widely applied to other types of problems without modification.
\end{itemize}

\section{Related Work}

\subsection{Dual Systems in Humans}
System 1 and System 2, or fast and slow thinking \citep{thinking,frankish2010dual,evans2008dual}, are common terms used in psychology. In the book $\textit{Thinking, Fast and Slow}$, System 1 refers to the brain's automatic and fast processes that operate with little effort and without a sense of conscious control. In contrast, System 2 focuses attention on mental activities that require effort, such as complex calculations. The operations of System 2 are often associated with the subjective experience of agency, choice, and concentration. Dual system theory is widely applied in psychological explanations of various behavioral phenomena in economic, social, and animal-conditioning contexts \cite{kahneman2002representativeness,loewenstein2004animal,killcross2002associative,dickinson2002role}.


Meanwhile, a broad range of neural and behavioral data suggests \citep{daw2005uncertainty} that the brain contains multiple systems for behavioral choice, including one associated with the prefrontal cortex and another with the dorsolateral striatum. This serves as one of the pieces of evidence that support the existence of dual systems as a real structure within the brain.
\subsection{Dual Systems Models}
Currently, LLMs also share many similarities with humans' System 1. As probabilistic models, they tend to output the most ``relevant'' answers to a given question and perform well when answering simple and familiar questions \citep{openai2023gpt}. However, when the questions become more complex, the logical reasoning abilities of large models are challenged. In fact, 
 \citet{chiang2023tighter} demonstrated that there are inherent limits to the problem-solving capabilities of large language models with a transformer architecture.

Surprisingly, simply prompting the model with ``let's think step by step'' enables it to break down complex problems into smaller steps, leading to a significant increase in accuracy. This type of approach, known as the Chain of Thought \citep{wei2022chain}, has become one of the most popular methods to improve the performance of LLMs. The effectiveness of Chain of Thought has been demonstrated by \citet{prystawski2024think,feng2024towards}. Next, we will categorize and introduce several System 2 models.
\subsubsection{The ``Chain of Thought'' Method}
Many improved versions have been developed based on the Chain of Thought approach. Tree of thoughts (ToT) \citep{yao2024tree} integrates the model’s abilities to produce and evaluate thoughts with search algorithms like breadth-first or depth-first search. Graph-of-thought \citep{yao2023beyond}, a graph-based framework advancing traditional sequential methods to better align with the non-linear characteristics of human thinking. For complex reasoning tasks with multiple valid paths, Self-Consistency Chain of Thought \citep{wangself} generates diverse reasoning chains by sampling from the language model’s decoder. It then identifies the most consistent final answer by marginalizing these sampled chains. Tree-of-mixed-thought \citep{hu2023tree} integrates one-stop reasoning (fast) with Tree-of-Thought (slow) and applies this combined approach to multi-hop visual reasoning tasks.

\subsubsection{Other Method}
In addition, many other prompt-based approaches have been proposed. For example, similar to Chain of Thought, the Divide-and-Conquer Program divides the entire task into smaller sub-tasks that can be processed in parallel \citep{zhang2024guiding}; prompts inspired by Aristotle's method of teaching students \citep{chang2023prompting}; TextGrad \citep{yuksekgonul2024textgrad}, which uses gradient descent on large model texts through prompts; Sys2 Attention \citep{weston2023system}, which directs LLMs to focus on important parts of the text through prompts; Take a Step Back \citep{zhengtake}, which extracts high-level abstract concepts; and the work of \citet{ma2023let} using counterfactuals to improve the moral reasoning of models .

Most of the methods mentioned above focus on a specific thinking mode that people use to solve problems, and some are tailored to specific types of problems or datasets, which limits their generalization. We aim to integrate multiple thinking modes and delegate the sub-tasks that arise during the thinking process to LLMs, thereby achieving better overall performance.In addition, we hope MTMT can also be applied in other fields in the future, such as RAG (Retrieval-Augmented Generation) \cite{shinn2024reflexion,yao2023react,asai2023self}.

\section{Methodology}

\subsection{Problem Definition}
Using the symbol $p$ to define the large language model (LLM) and $Q$ as the question being asked, the MTMT is represented as $G_{p}$, with the initialization as follows:
\begin{equation}
    G_{p}(Q) = q_{1},
\end{equation}
\begin{equation}
    P(q_{1}) = a_{1},
\end{equation}
where $q_{1}$ represents the first sub-question derived from $G_{p}$ regarding $Q$, and $a_{1}$ is the answer provided by the LLM to the first sub-question.

For the subsequent $i$-th step, the iteration proceeds as follows:
\begin{equation}
    G_{p}(Q, a_{1}, a_{2}, \dots, a_{i}) = q_{i+1} \label{eq1},
\end{equation}
\begin{equation}
    P(q_{i+1}) = a_{i+1} \label{eq2}.
\end{equation}
This represents the repeated ``communication'' between System 1 and System 2.
Based on the answers from the previous $i$ steps and the question $Q$, we derive the $q_{i+1}$ and interact with the large model to obtain the answer $a_{i+1}$ (note that the $q_{i+1}$ includes the information needed for the LLM to answer it).

Each time we complete a sub-question directly connected to $Q$, we obtain an answer based on $Q$ and $a_{1}, a_{2}, \dots, a_{j}$:
\begin{equation}
    G_{p}(Q, a_{1}, a_{2}, \dots, a_{j}) = A_{k},
\end{equation}
where $A_{k}$ represents the answer generated for $Q$ during the $k$-th iteration.
Once certain criteria are met (see Section \ref{node}), $A_{k}$ will be considered as the final answer.
During the execution of this algorithm, we ensure that the sub-question $q_{i}$ is intuitive, simple, and suitable for System 1 (LLM) to answer. We then perform information extraction, evaluation, modification, and filtering on the obtained answers $a_{i}$ before applying them to the original question $Q$.

The overall process for Equations (\ref{eq1}) and (\ref{eq2}) is as follows (also shown in Figures \ref{framework}) :

\begin{enumerate}
\item Based on different thinking mode sub-questions, and considering the known pairs $(q_{i}, a_{i})$, generate the appropriate prompt and obtain the model's response.
\item We will re-engage the model to perform information extraction in a dictionary format (see Appendix \ref{appendix:information extraction} for more details). This process generates nodes that store the extracted information (for logical processing) and the original response (for further model generation), which are then added to the graph.
\item Select the next thinking mode type to be used and transition to the designated node.
\end{enumerate} 
Notably, we do not require the LLM to output results in a specific format on the first response. Instead, we ask it to extract the relevant information after its initial answer based on the content provided. This is because discouraging prompts, such as “just tell me the result without any explanation,” can negatively affect the LLM's reasoning ability \cite{zhao-etal-2024-enhancing-zero,zhou2023complex}.

\subsection{Thinking Mode}
Specifically, we categorize thinking modes into the following major types: \textbf{decompose}, \textbf{association}, \textbf{compare}, \textbf{importance}, \textbf{inference} and others. Each major type contains several different prompts. See Appendix \ref{appendix:thinking mode prompt} for specific details about each prompt. The use of different thinking modes serves two main purposes: generating thinking nodes and performing operations on different nodes.
\subsubsection{Generating Thinking Nodes}
By generating prompts based on the thinking modes, we can obtain a wealth of useful information.
For example, consider the following question:

\textit{Question:There are only three people on the playground Xiao Ming, Xiao Hong, and Xiao Li. Xiao Ming is running, Xiao Hong is playing tennis with someone, so what is Xiao Li doing?}

We can inquire about the question type:

\textit{What is the task mentioned above?specify the type of task (e.g., mathematics, biology, general knowledge and so on).}

And obtain the model's response:

\textit{The task mentioned above is a logic puzzle or a riddle.}

This information can help LLM address $Q$ and $a_{i}$. Additionally, the importance thinking mode can assist in focusing on the critical information within $a_{i}$. The decomposition thinking mode breaks down the target problem into several steps. The association thinking mode helps us find a series of related pieces of information based on the given data.
\subsubsection{Performing Operations on Nodes}
Thinking modes not only generate a wealth of information for creating thinking nodes but also influence the entire graph with various prompts. For example, in the importance category,  \textit{unimportant\_point} helps us select useful $a_{i}$ data. In the decompose category, \textit{decompose\_task} determines whether to proceed with task decomposition and how many steps to break it down into.


\subsection{Thinking Node}
\label{node}
The question $Q$ serves as the root node. For generating subsequent nodes, the perplexity is used to measure whether the model is ``confident'' or ``confused'' about the question. In LLMs, for a given response $S = (t_{1}, t_{2}, \dots, t_{N})$, where $t_{i}$ represents the $i$-th token and $N$ represents the total number of tokens, the perplexity is calculated as:
\begin{equation}
   PP(S) = \sqrt[N]{\prod_{i=1}^{N} \frac{1}{P\left(t_{i} \mid t_{1} \ldots t_{i-1}\right)}}.
\end{equation}
The generation, regeneration and deactivation of other nodes follow these guidelines:

\subsubsection{Node Generation}
Each node has a perplexity threshold, which is calculated as follows:
\begin{equation}
    PPT_{i} = PPT_{0}+\alpha D(i) ,
\end{equation}
where $PPT_{i}$ is the perplexity threshold of the i-th node, $PPT_{0}$ is the initial perplexity threshold, $\alpha$ is the proportionality coefficient, and $D(i)$ represents the number of nodes in the shortest path from node $i$ to the root node (excluding the root node).

By calculating perplexity, if the perplexity of a given response exceeds a certain threshold, we will continue generating sub-nodes using other types of strategies for that node's question to obtain more information and methods until the perplexity requirement is met. If a node and its parent node both exhibit ``confusion'', a breadth-first search (BFS) approach is adopted, prioritizing further exploration of the parent node using different strategies to extract more information.

For selecting a thinking mode, if a specific mode has already been assigned by a previous thinking mode, we follow the assigned strategy. Otherwise, a thinking mode will be randomly chosen from all available modes to generate the next node. Initially, we always let the MTMT go through \textit{task\_recognition} and \textit{decompose\_task}.

\subsubsection{Node Regeneration}
After generating information for other sub-nodes, we regenerate the node. This process is repeated until the perplexity requirement is satisfied. Once the perplexity condition is met, we use the \textit{difference\_answer} in compare category to compare the quality of the answers generated in both instances and ultimately select the most suitable response.
\subsubsection{Node Deactivation}
Not all information stored in each node is always useful; irrelevant or incorrect information can even reduce the model's accuracy. For nodes that have met the perplexity requirement, we use the \textit{unimportant\_point} in importance category to assess whether the information from the sub-nodes contributes to resolving the parent node's question. If it does, we incorporate this information into the prompt used for regenerating the parent node.

\section{Experiments}

\subsection{Base Model}
We will use OpenAI's latest large language model, GPT-4o mini \citep{GPT-4o-mini}, as the System 1. GPT-4o mini enables a broad range of tasks with its low cost and latency, and it performs well in extracting structured data, making it an ideal base model for our experiments.
\subsection{Datasets}
We will test the model's performance on the following three datasets.

\textbf{GSM8K \citep{patel-etal-2021-nlp}:}
The dataset contains math word problems geared toward an average middle-school curriculum, which is also repeatedly adopted by prior work as a key benchmark for arithmetic reasoning. We use a total of 1,319 data points from its test set.

\textbf{GPQA \citep{rein2023gpqa}:} 
Google-proof Question Answering (GPQA) is a recent benchmark where challenging multiplechoice questions in physics, biology, and chemistry are created and labeled by domain experts who have or are pursuing PhD degrees. In this benchmark, experts and skilled non-experts are reported to achieve 81$\%$ and 22$\%$ accuracy respectively, demonstrating the difficulty of the questions. We use a total of 448 data points from its dataset.

\textbf{TruthfulQA \citep{TruthfulQA}:} 
TruthfulQA is a classification (multi-choice) task designed to test LLMs’ propensity to dispense harmful information. The dataset contains 654 test instances. We use a total of 817 multiple-choice questions from the dataset.

\subsection{Baseline}
Here are the baselines for comparison. These baselines are widely used to evaluate the accuracy of LLMs. See Appendix \ref{appendix:baseline prompt} for specific prompts.

\textbf{GPT-4o mini:} 
Using the original model directly on the dataset.

\textbf{GPT-4o mini + CoT:}
GPT-4o mini model is queried with zero-shot CoT prompting \cite{zeroshotcot}: ``Let’s think step by step'' is appended to the question.

\textbf{GPT-4o mini + 3-shot:}Three examples with corresponding answers are added to the prompt input of the GPT-4o mini model.

\textbf{GPT-4o mini + CoT 1-shot:}For 1-shot, One demonstration example of a question and answer pair is provided in the prompt, where the answer is in the style of CoT \cite{wei2022chain}.

\subsection{Result and Compare}
The results of different methods on various datasets are shown in Table \ref{table}.

\begin{table}[htbp]
\small
  \centering
    \begin{tabularx}{\columnwidth}{X|X|X|c|c|c}  
    \hline
    \multicolumn{3}{X|}{\textbf{Method}} & \textbf{GPQA} & \textbf{TruthfulQA} & \textbf{GSM8K} \bigstrut\\
    \hline
    \multicolumn{3}{X|}{GPT-4o mini } & 38.8\% & 55.4\% & 93.6\% \bigstrut[t]\\
    \multicolumn{3}{r|}{+ CoT} & 39.1\% & 57.6\% & 93.3\% \\
    \multicolumn{3}{r|}{+ 3-shot} & 41.1\% & \textbf{58.9\%} & 93.1\% \\
    \multicolumn{3}{r|}{+ CoT 1-shot} & 40.2\% & 57.3\% & \textbf{93.9\%} \\
    \multicolumn{3}{r|}{+ MTMT} & \textbf{44.0\%} & \textbf{58.5\%} & \textbf{93.9\%} \bigstrut[b]\\
    \hline
    \end{tabularx}%
   \caption{Comparison between the MTMT and other baselines. The MTMT shows a 5.2\% improvement over the base model on the GPQA dataset.}     
  \label{table}%
\end{table}%

\begin{figure*}[ht]
    \centering
    \includegraphics[width=0.95\textwidth,trim={1.5cm 0.1cm 1.5cm 0.1cm},clip]{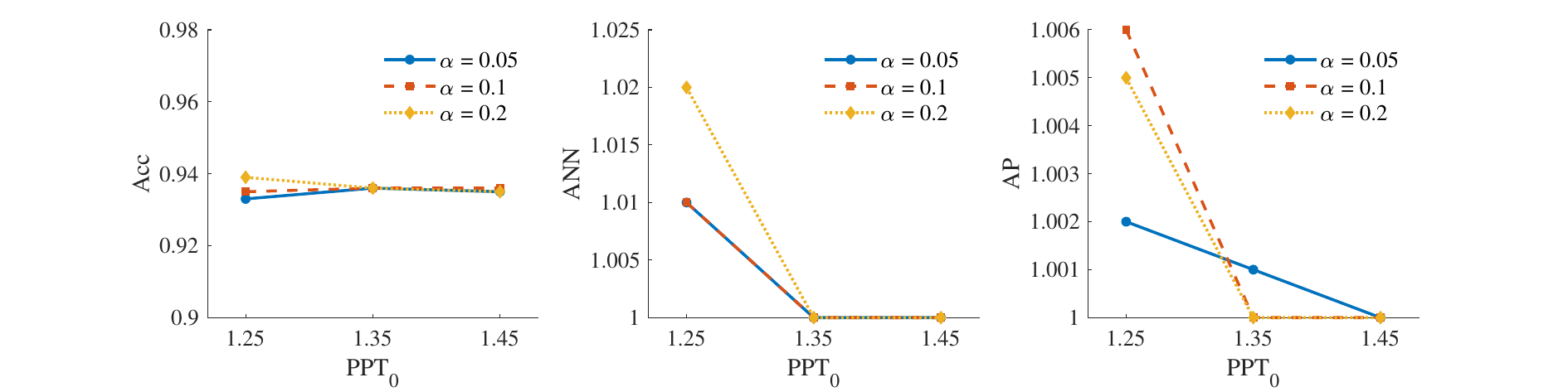}
    \caption{The Impact of $PPT_{0}$ and $\alpha$ on $Acc$, $ANN$, and $AP$ in the GSM8K Dataset.}
    \label{polylineGSM8K}
\end{figure*}

From the table, we can see that our method shows a 5.2\% and 3.1\% improvement on the TruthfulQA and GPQA datasets, respectively, compared to direct LLM responses. However, the difference on the GSM8K dataset is negligible, likely because GSM8K mainly involves middle school-level math problems with lower complexity. The base model has already achieved good results on GSM8K and is confident in its responses, as shown in Figure \ref{polylineGSM8K}, where almost no additional thinking nodes were generated.

\begin{figure*}[ht]
    \centering
    \includegraphics[width=0.95\textwidth,trim={1.5cm 0.1cm 1.5cm 0.1cm},clip]{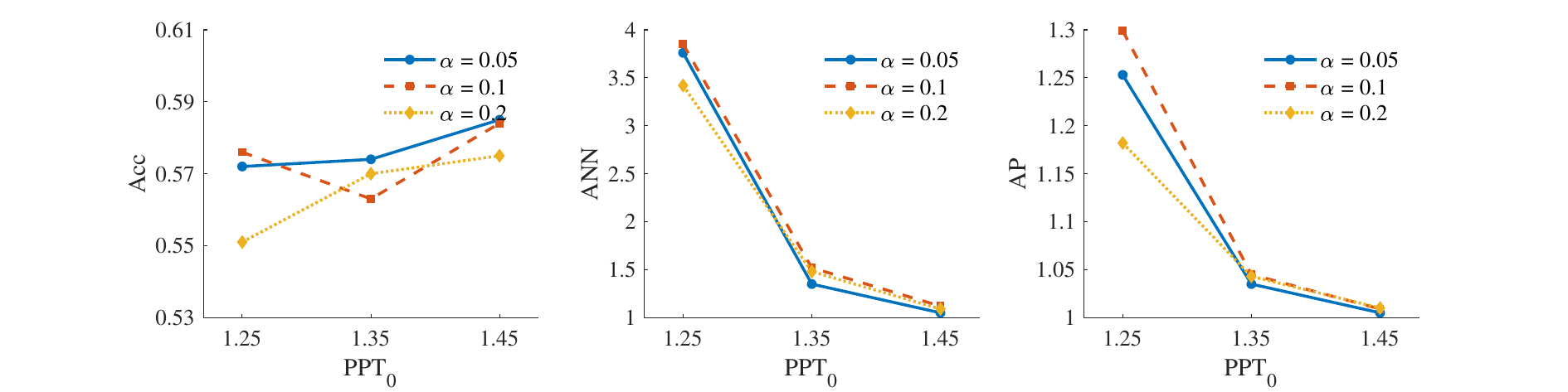}
    \caption{The Impact of $PPT_{0}$ and $\alpha$ on $Acc$, $ANN$, and $AP$ in the TruthfulQA Dataset.}
    \label{polylineTruthfulQA}
\end{figure*}

\begin{figure*}[ht]
    \centering
    \includegraphics[width=0.95\textwidth,trim={1.5cm 0.1cm 1.5cm 0.1cm},clip]{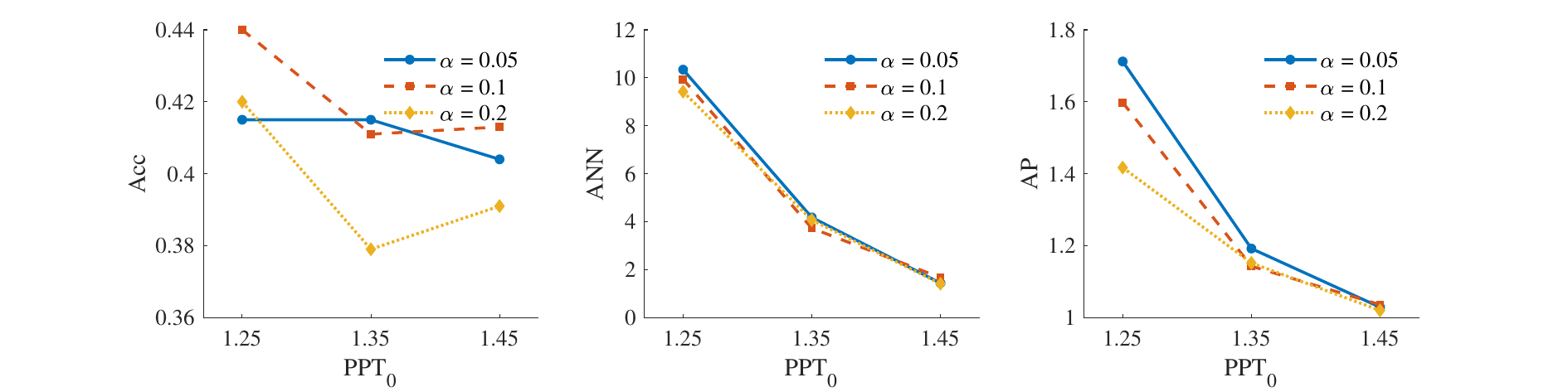}
    \caption{The Impact of $PPT_{0}$ and $\alpha$ on $Acc$, $ANN$, and $AP$ in the GPQA Dataset.}
    \label{polylineGPQA}
\end{figure*}

We then tested the impact of different parameters on MTMT. Considering the API computational limitations, we set the maximum number of generated thinking nodes to 30.The results 
 are shown in Figure \ref{polylineTruthfulQA}, Figure \ref{polylineGSM8K} and Figure \ref{polylineGPQA}.
 
 Here, we denote $ANN$ as the average number of nodes and $AP$ as the depth of the graph, with the following calculation formula:
\begin{equation}
   AP = \sum^{n}_{j=0} (max(D(j_{1}),D(j_{2}),\dots,D(j_{k}))+1), 
\end{equation}
where $j$ represents the $j$-th data point in the dataset, and $j_{k}$ represents the $k$-th thinking node generated for the $j$-th data point. The line graph shows that, in terms of accuracy, the accuracy of GPQA declines as $PPT_{0}$ increases, while TruthfulQA shows an upward trend. This aligns with the intuitive difficulty of the two datasets—GPQA presents more challenging tasks, requiring the generation of more thinking nodes. Additionally, the choice of $\alpha$ does not significantly impact accuracy, likely due to our maximum limit of 30 thinking nodes and the breadth-first strategy, preventing MTMT from generating overly deep thoughts.

For both $ANN$ and $AP$, as $PPT_{0}$ increases, the values continuously decrease, indicating a reduction in the number of generated thinking nodes and the depth of the graph. Additionally, the value of $\alpha$ also affects the graph's depth; a smaller $\alpha$ results in a deeper graph.

\begin{figure}[h]
    \centering
    \includegraphics[width=0.45\textwidth]{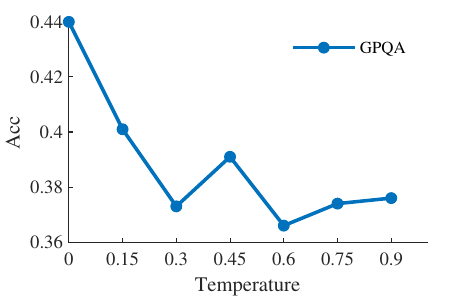}
    \caption{Relationship between accuracy and LLM temperature on the GPQA dataset. As the temperature continues to increase, the accuracy decreases correspondingly. Therefore, in the other experiments in this paper, we set the default temperature to 0.}
    \label{temperature}
\end{figure}

At the same time, we also explored the impact of the temperature parameter in LLMs on the accuracy of the MTMT model (with the perplexity threshold set at 1.25). The experimental results are shown in Figure \ref{temperature}.

From the figure, we can observe that as the temperature of the LLM increases, its ``creativity'' also rises, but the accuracy significantly decreases, even yielding worse results compared to the model's direct output. Moreover, further increasing the temperature leads to the model's inability to extract relevant information in the specified format. Therefore, in other experiments, we consistently set the default temperature to 0.

\subsection{Ablation Study}
To evaluate the effectiveness of different types of thinking modes, we conducted ablation experiments on the GPQA and TruthfulQA datasets. In these experiments, we remove one specific thinking mode and test its impact on accuracy. We conducted ablation experiments on the GPQA dataset under the conditions of $PPT_{0} = 1.25$ and $\alpha = 0.1$, and on the TruthfulQA dataset under the conditions of $PPT_{0} = 1.45$ and $\alpha = 0.05$. The results are shown in Table \ref{GPQA ablation} and Table \ref{TruthfulQA ablation}.

\begin{table}[htbp]
  \centering
    \begin{tabular}{l|r}
    \multicolumn{2}{c}{\textbf{GPQA Acc: 44.0\%}} \\
    \midrule
    Removed thinking mode & \multicolumn{1}{l}{Accuracy} \\
    \midrule
    Decompose &  39.3\% ~~4.7\% $\downarrow$\\
    \midrule
    Association &  39.5\% ~~4.5\% $\downarrow$ \\
    \midrule
    Compare &  42.6\% ~~1.4\% $\downarrow$\\
    \midrule
    Importance &  41.7\% ~~2.3\% $\downarrow$\\
    \midrule
    Inference &  41.3\% ~~2.7\% $\downarrow$\\
    \end{tabular}%
  \caption{The ablation experiment on the GPQA dataset. Both the decompose mode and the association mode have a significant impact on accuracy.}    
  \label{GPQA ablation}%
\end{table}%

\begin{table}[htbp]
  \centering
    \begin{tabular}{l|r}
    \multicolumn{2}{c}{\textbf{TruthfulQA Acc: 58.5\%}} \\
    \midrule
    Removed thinking mode & \multicolumn{1}{l}{Accuracy} \\
    \midrule
    Decompose &  56.4\% ~~2.1\% $\downarrow$ \\
    \midrule
    Association &  57.9\% ~~0.6\% $\downarrow$ \\
    \midrule
    Compare &  57.2\% ~~1.3\% $\downarrow$ \\
    \midrule
    Importance &  57.6\% ~~0.9\% $\downarrow$\\
    \midrule
    Inference &  57.3\% ~~1.2\% $\downarrow$\\
    \end{tabular}%
  \caption{The ablation experiment on the TruthfulQA dataset. The removal of each mode did not result in a significant performance drop, with the decompose mode causing the largest decrease.}    
  \label{TruthfulQA ablation}%
\end{table}%

From the table, we can observe that removing different thinking modes has varying impacts on the results. The removal of the decompose mode has the most significant effect, leading to a 4.7\% and 2.1\% decrease in accuracy on the GPQA and TruthfulQA datasets, respectively. On the GPQA dataset, the removal of the association mode also resulted in a 4.5\% decrease in accuracy.Other thinking modes show similar effects, causing accuracy drops ranging from 2.7\% to 0.6\%
\subsection{Analysis}
MTMT has enhanced the LLM's logical reasoning and explainability in various instances. For example, in response to the question, \textit{``What are the richest countries in South America by GDP per capita?''}, MTMT prompted the model to identify the type of question and the methods required to solve it (such as understanding GDP per capita and identifying the countries). This interaction enabled the model to conclude, \textit{``It's important to note that while Venezuela has historically had a high GDP per capita due to its oil wealth, its current economic situation has significantly affected its GDP per capita ranking.''} This corrected an error found in the baseline output.
Such instances are not isolated within this approach, indicating that we can achieve further advancements in the LLM's controllable output, logical reasoning, and explainability.

The errors in MTMT can be categorized into three main types: \textbf{overly difficult tasks}, \textbf{information accumulation}, and \textbf{base model issues}. For instance, in the GPQA dataset, the prevalence of advanced mathematics increases computational demands. Given our computational limits, we restrict the generation of nodes to a maximum of 30 for each question, which may not be sufficient for the number of nodes needed to solve such complex problems. 

Information accumulation is another challenge. As different nodes are generated, the amount of information sent to the base model increases, leading to confusion about which points to focus on. Summarizing and refining information may help address this issue. 

Additionally, many errors stem from the underlying model. For example, in certain tasks, the model fails to produce output in the required format, rendering it ineffective. In simpler questions, such as comparing 3.8 and 3.11, the model can also provide incorrect answers. Such errors are likely to accumulate during the node generation process in MTMT. However, we believe that as LLMs continue to develop, these types of issues will diminish.

\section{Conclusion}
Recently, many studies have focused on designing prompts based on specific thought processes and repeatedly engaging large language models (LLMs), achieving remarkable results. This paper introduces a method called MTMT, which aggregates various human problem-solving approaches to enable LLM to provide more accurate answers. This multi-faceted thinking approach offers the model additional background knowledge and guidance, breaks down complex tasks into sub-questions that are easier to solve accurately. Across three different datasets, this method outperformed models using Chain of Thought (CoT). Moreover, we further investigated the impact of different parameters on MTMT's accuracy and conducted ablation experiments on various thinking modes. Overall, MTMT can enhance the LLM's ability to solve complex problems.
\section{Limitations}
Much relevant information may affect the model's performance. Overly long prompts might prevent the base model from focusing on the question to be answered. In future work, incorporating some concise methods in the thinking mode could help address this issue.
Also, search methods like MTMT require more resources (e.g., GPT-4o mini API costs) than few-shot learning to improve task performance. However, adjusting the perplexity threshold allows users to customize the trade-off between performance and cost. As more efficient smaller models are introduced in the future, these costs are expected to decrease. 
Finally, future work could consider integrating a memory module into the network, enabling it to utilize relevant temporary memories to answer related questions, thereby enhancing the model's safety and long-term memory capabilities.

\bibliography{custom}

\appendix

\section{Thinking Mode Prompt Template}
\label{appendix:thinking mode prompt}
The use of different thinking modes serves two main purposes: generating thinking nodes and performing operations on different nodes. Table \ref{thinking mode} shows the prompts of different thinking modes. 
\begin{table*}[htbp]
  \centering

    \begin{tabularx}{\textwidth}{p{8em}|c|p{8em}|c|c|c}  
    \multicolumn{6}{X}{\textbf{Decompose}} \\
    \midrule
    \multicolumn{2}{p{8em}|}{Decompose\_task} & \multicolumn{4}{X}{Please break down the <question>\{problem\}</question> into several steps and briefly describe the work that should be done in each step. Note that you do not need to provide the answer for each step at this time.} \\
    \midrule
    \multicolumn{2}{p{8em}|}{Step} & \multicolumn{4}{X}{For the step\{i\}, what should we do? Please provide an answer based on the issue, the answers to the previous steps, and the goal of step\{i\}.} \\
    \midrule
    \multicolumn{6}{X}{\textbf{Association}} \\
    \midrule
    \multicolumn{2}{p{8em}|}{Association} & \multicolumn{4}{X}{What words/stories/rules/theorems does <item>\{item\}</item> remind us of? Please explain each one.} \\
    \midrule
    \multicolumn{2}{p{8em}|}{Similar\_problem} & \multicolumn{4}{X}{What other problems are similar to <question>\{problem\}</question>? Please provide an example and its solution.} \\
    \midrule
    \multicolumn{6}{X}{\textbf{Compare}} \\
    \midrule
    \multicolumn{2}{p{8em}|}{Compare} & \multicolumn{4}{X}{What are the similarities and differences between <thing1>\{thing1\}</thing1> and <thing2>\{thing2\}</thing2>? Please provide a detailed explanation and answer in separate parts} \\
    \midrule
    \multicolumn{2}{p{8em}|}{Compare\_ordinary} & \multicolumn{4}{X}{Compared to the usual tasks, what are the differences in this <question>\{problem\}</question>? Please answer in separate parts.} \\
    \midrule
    \multicolumn{2}{p{8em}|}{Difference\_impact} & \multicolumn{4}{X}{What impact do these <differences>\{differences\}</differences> have on the <question>\{problem\}</question>?} \\
    \midrule
    \multicolumn{2}{p{8em}|}{Difference\_answer} & \multicolumn{4}{X}{For the specific <question>\{problem\}</question>, What are the differences between <answer1>\{answer1\}</answer1> and <answer2>\{answer2\}?} \\
    \midrule
    \multicolumn{2}{p{8em}|}{Choose\_answer} & \multicolumn{4}{X}{Which answer is better under this <question>\{problem\}</question>? answer1:{answer1} \textbackslash{}n answer2:\{answer2\} \textbackslash{}n The diferences between two answers is <differences>\{differences\}</differences>. Please provide your reasons.Finally, choose the better one.} \\
    \midrule
    \multicolumn{6}{X}{\textbf{Importance}} \\
    \midrule
    \multicolumn{2}{p{8em}|}{Importance} & \multicolumn{4}{X}{What is the most important aspect of <question>\{problem\}</question>?} \\
    \midrule
    \multicolumn{2}{p{8em}|}{Unimportant\_point} & \multicolumn{4}{X}{The following text is the relative point to the problem. Please select some unimportant or irrelevant nodes in solving the problem.The problem is <question>\{problem\}</question>.The nodes text is <node\_text>{node\_text}</node\_text>.} \\
    \midrule
    \multicolumn{2}{p{8em}|}{ Help\_judgment} & \multicolumn{4}{X}{This <item>\{item\}</item> is related to <main\_thing> \{main\_thing\} </main\_thing>. Is it helpful in solving the <question>\{problem\}</question>?} \\
    \midrule
    \multicolumn{6}{X}{\textbf{Inference}} \\
    \midrule
    \multicolumn{2}{p{8em}|}{Counter\_factual1} & \multicolumn{4}{X}{If <thing>\{thing\}</thing> does not exist, what impact would it have on the <result>\{result\}</result>?} \\
    \midrule
    \multicolumn{2}{p{8em}|}{Counter\_factual2} & \multicolumn{4}{X}{If <thing>\{thing\}</thing> is opposite, what impact would it have on the <result>\{result\}</result>?} \\
    \midrule
    \multicolumn{2}{p{8em}|}{Reason} & \multicolumn{4}{X}{What is the reason for <thing>\{thing\}</thing> occurring?} \\
    \midrule
    \multicolumn{2}{p{8em}|}{Result} & \multicolumn{4}{X}{What kind of impact or outcome will this <thing>\{thing\}</thing> bring?} \\
    \midrule
    \multicolumn{2}{p{8em}|}{Define} & \multicolumn{4}{X}{What is the definition of <thing>\{thing\}</thing>?} \\
    \end{tabularx}%
    \caption{Thinking Mode Prompt Template}
  \label{thinking mode}%
\end{table*}%

\section{Information Extraction  Prompt Template}
\label{appendix:information extraction}
We also need to extract relevant information from the LLM's responses to perform additional logical operation and the final answer (e.g., multiple-choice options like A, B, C, D or numerical answers) also needs to be extracted using this template (see Table \ref{extract}).

\begin{table*}[htb]
  \centering
    \begin{tabularx}{\textwidth}{X X} 
    \multicolumn{2}{X}{\textbf{Extract Information}} \\
    \midrule
    \multicolumn{2}{X}{Next, a segment of Q\&A text will be provided. Please extract information according to the following format.} \\
    \multicolumn{2}{X}{chatting records = \{text\}} \\
    \multicolumn{2}{X}{\{format\_instructions\}} \\
    \end{tabularx}%
    \caption{Extract Information from LLM's Response}    
  \label{extract}%
\end{table*}%

The format\_instructions are generated from the Langchain library (version: 0.2.3), an open-source software library. For a set of attributes to be extracted, given their descriptions, the generated format instructions are shown in Table \ref{format instructions}.

\begin{table*}[htbp]
  \centering
    \begin{tabularx}{\textwidth}{X X} 
    \multicolumn{2}{X}{\textbf{format\_instructions}} \\
    \midrule
    \multicolumn{2}{X}{The output should be a markdown code snippet formatted in the following schema, including the leading and trailing "```json" and "```":\newline{}\newline{}```json\newline{}\{\newline{}	"object 1": string  // description 1\newline{}	"object 2": bool  // description 2.\newline{}	"object 3": int  // description 3\newline{}\}\newline{}```} \\
    \end{tabularx}%
  \caption{Format Instructions}    
  \label{format instructions}%
\end{table*}%
\begin{table*}[htbp]
  \centering
    \begin{tabularx}{\textwidth}{p{8em}|c|p{8em}|X}  
    \multicolumn{3}{l|}{\textbf{Baseline}} & \textbf{Prompt} \\
    \midrule
    \multicolumn{3}{l|}{GPT-4o mini} & Question: {question}\newline{}Answer: \\
    \midrule
    \multicolumn{3}{l|}{GPT-4o mini + CoT} & Question: {question}\newline{}\newline{}Let's think step by step.\newline{}Answer: \\
    \midrule
    \multicolumn{3}{l|}{GPT-4o mini + 3-shot} & Question: {question}\newline{}\newline{}Here are few examples:\newline{}{few\_shots}\newline{}\newline{}Answer: \\
    \midrule
    \multicolumn{3}{l|}{GPT-4o mini + CoT 1-shot} & Question: {question}\newline{}\newline{}Here is a step-by-step example:\newline{}{shot}\newline{}\newline{}Answer: \\
    \end{tabularx}%
    \caption{Baseline Prompt Template}
  \label{baseline prompt}%
\end{table*}%
Table \ref{object_description} represents various objects and their corresponding descriptions.

\begin{table*}[htbp]
  \centering
    \begin{tabularx}{\textwidth}{p{8em}|c|p{8em}|c|c|c}
    \multicolumn{2}{p{9em}|}{\textbf{Object}} & \multicolumn{4}{X}{\textbf{Description}} \\
    \midrule
    \multicolumn{2}{p{9em}|}{type of task} & \multicolumn{4}{X}{Based on the reply, specify the type of task.} \\
    \midrule
    \multicolumn{2}{p{9em}|}{approach and considerations} & \multicolumn{4}{X}{The approach and key considerations about this task.} \\
    \midrule
    \multicolumn{2}{p{9em}|}{number\_step} & \multicolumn{4}{X}{Into how many steps can the question be broken down? Just give a number.} \\
    \midrule
    \multicolumn{2}{p{9em}|}{the solution of step\{i\}} & \multicolumn{4}{X}{The solution of step{i}} \\
    \midrule
    \multicolumn{2}{p{9em}|}{number\_associate\_item} & \multicolumn{4}{X}{How many items are mentioned related to the main item? Just give a number.} \\
    \midrule
    \multicolumn{2}{p{9em}|}{item\{i\}} & \multicolumn{4}{X}{Write the name of item{i} in order.} \\
    \midrule
    \multicolumn{2}{p{9em}|}{example} & \multicolumn{4}{X}{The example of similar question} \\
    \midrule
    \multicolumn{2}{p{9em}|}{answer} & \multicolumn{4}{X}{The correspond answer} \\
    \midrule
    \multicolumn{2}{p{9em}|}{similarities} & \multicolumn{4}{X}{The similarities between two things} \\
    \midrule
    \multicolumn{2}{p{9em}|}{differences} & \multicolumn{4}{X}{The differences between two things} \\
    \midrule
    \multicolumn{2}{p{9em}|}{differences} & \multicolumn{4}{X}{The differences between usual tasks and this specific task} \\
    \midrule
    \multicolumn{2}{p{9em}|}{impact} & \multicolumn{4}{X}{What impact do these differences have on the problem} \\
    \midrule
    \multicolumn{2}{p{9em}|}{differences} & \multicolumn{4}{X}{The differences between answer1 and answer2} \\
    \midrule
    \multicolumn{2}{p{9em}|}{better\_answer} & \multicolumn{4}{X}{Which answer is better under this question? Answer answer1 if answer1 is better, answer2 if answer2 is better.} \\
    \midrule
    \multicolumn{2}{p{9em}|}{important\_item} & \multicolumn{4}{X}{What is the most important item in this question?} \\
    \midrule
    \multicolumn{2}{p{9em}|}{irrelevant\_point} & \multicolumn{4}{X}{What are irrelevant nodes in the answer? Just give the number of the node.} \\
    \midrule
    \multicolumn{2}{p{9em}|}{judgment} & \multicolumn{4}{X}{Based on the reply,is it helpful in solving the problem? Answer True if yes, False if not or unknown.} \\
    \midrule
    \multicolumn{2}{p{9em}|}{reason} & \multicolumn{4}{X}{The reasons mentioned in the answer} \\
    \midrule
    \multicolumn{2}{p{9em}|}{definition} & \multicolumn{4}{X}{The definition mentioned in the answer} \\
    \midrule
    \multicolumn{2}{p{9em}|}{impact\_or\_outcome} & \multicolumn{4}{X}{The impact or outcome mentioned in the answer} \\
    \midrule
    \multicolumn{2}{p{9em}|}{impact} & \multicolumn{4}{X}{The impact mentioned in the answer} \\
    \end{tabularx}%
 \caption{Objects and their corresponding descriptions, used for information extraction.}
  \label{object_description}%
\end{table*}%
\section{Baseline Prompt Template}
\label{appendix:baseline prompt}
And here are the prompt templates of baseline (see Table \ref{baseline prompt}).

\end{document}